\documentclass{article}
\usepackage{amsfonts}
\usepackage{microtype}
\usepackage{graphicx}
\usepackage{subfigure}
\usepackage{booktabs} % for professional tables
\usepackage{xspace}
\usepackage{multirow}
\usepackage{amsmath} 
\usepackage{threeparttable}
\usepackage{tabularx} 
\usepackage{hyperref}
\usepackage{comment}
\newcommand{\name}{AttnCache\xspace}
\usepackage[accepted]{mlsys2025style/mlsys2025}

\mlsystitlerunning{\name: Accelerating Self-Attention Inference for LLM Prefill via Attention Cache}

\begin{document}
% \twocolumn[
% \mlsystitle{\name: Accelerating Self-Attention Inference for LLM Prefill via Attention Cache}
% \maketitle
% \vskip 0.3in
% ]

\twocolumn[

% \mlsystitle{\name: Fast Self-Attention Inference for LLM Prefill via Attention Cache}

\mlsystitle{\name: Accelerating Self-Attention Inference for LLM Prefill via Attention Cache}

\mlsyssetsymbol{equal}{*}

\begin{mlsysauthorlist}
\mlsysauthor{Dinghong Song}{ucm}
\mlsysauthor{Yuan Feng}{ucm}
\mlsysauthor{Yiwei Wang}{ucm}
\mlsysauthor{Shangye Chen}{ucm}
\mlsysauthor{Cyril Guyot}{wdc}
\mlsysauthor{Filip Blagojevic}{wdc}
\mlsysauthor{Hyeran Jeon}{ucm}
\mlsysauthor{Pengfei Su}{ucm}
\mlsysauthor{Dong Li}{ucm2}
\end{mlsysauthorlist}

\mlsysaffiliation{ucm}{University of California, Merced, Merced, CA, USA}
\mlsysaffiliation{ucm2}{Yotta Labs \& University of California, Merced, Merced, CA, USA}
\mlsysaffiliation{wdc}{Western Digital Research, San Jose, CA, USA}

\mlsyscorrespondingauthor{Dinghong Song}{dsong15@ucmerced.edu}
% \mlsyscorrespondingauthor{Dong Li}{dli35@ucmerced.edu}

%%%%%%%%%%%%%%%%%%%%%%%%%%%%%%%%

% You may provide any keywords that you
% find helpful for describing your paper; these are used to populate
% the "keywords" metadata in the PDF but will not be shown in the document
\mlsyskeywords{Machine Learning, MLSys}

\vskip 0.3in

\begin{abstract}
% Self attention, the kernel of transformer models, is computationally intensive, and hence a major performance-optimization focus for accelerating large language model (LLM) inference. Each request in LLM inference goes through a prefill phase for context encoding and a decoding stage for token generation.
% In this paper, we focus on using LLMs as text encoders to generate high-quality sentence representations and introduce \name, an innovative approach to accelerating self-attention inference in the prefill phase. \name draws inspiration from the intriguing observation of recurring and rich similarities in attention map computation across different inference sequences. Based on a memorization database that leverages emerging big memory systems, we propose efficient caching and searching techniques to identify similar attention maps and reuse them during inference, thereby reducing self-attention computation overhead.
% Our experiments are conducted on sentence embedding generation tasks from STS, SST-2, and MMLU. The experimental results show that \name achieves an average of 1.2× end-to-end speedup and 2× attention speedup on CPU, and 1.6× end-to-end speedup and 3× attention speedup on GPU, while incurring only negligible performance degradation. AttnCache is Github-available: https://anonymized.

% %%%%%%%%%%%%%%%%%%%%%%%%

Large Language Models (LLMs) are widely used in generative applications such as chatting, code generation, and reasoning. However, many real-world workloads—such as classification, question answering, recommendation, and text embedding—rely solely on the prefill stage of inference, where the model encodes input sequences without performing autoregressive decoding. In these prefill-only scenarios, the self-attention computation becomes the primary performance bottleneck due to its quadratic complexity with respect to sequence length. In this paper, we observe that semantically different sentences often produce similar attention maps across layers and heads. Building on this insight, we propose AttnCache, a framework that accelerates the prefill stage of LLM inference by retrieving and reusing similar attention maps. Based on an attention map memoization database, AttnCache employs efficient caching and similarity search techniques to identify and reuse pre-cached attention maps during inference, thereby reducing the computational overhead of self-attention.
Experimental results show that \name achieves an average of 1.2$\times$ end-to-end and 2$\times$ attention speedup on CPU, and 1.6$\times$ end-to-end and 3$\times$ attention speedup on GPU, with negligible accuracy degradation. AttnCache is available at: 
% \href{https://github.com/dinghongsong/AttnCache}
% {https://github.com/dinghongsong/AttnCache}.
\href{https://github.com/dinghongsong/AttnCache}
{https://github.com/PASAUCMerced/AttnCache}

% Based on a memorization database that leverages emerging big memory systems, we propose
% efficient caching and searching techniques to identify similar attention maps and reuse them during inference,
% thereby reducing self-attention computation overhead. 

% Building on this insight, we propose AttnCache, a framework that accelerates the prefill stage of LLM inference by retrieving and reusing precomputed attention maps from a lightweight attention map database. 

% AttnCache introduces (1) a compact neural embedding model to efficiently represent attention maps for similarity search, and (2) a memory-mapped storage mechanism that minimizes data copying and improves spatial and temporal locality across attention map accesses.

% We further demonstrate that AttnCache effectively extends to CPU environments, where memory capacity is larger and throughput is often more critical than latency. 

% These results highlight AttnCache as a practical and effective system for accelerating the prefill stage of LLM inference.

\end{abstract}
]

% this must go after the closing bracket ] following \twocolumn[ ...

% This command actually creates the footnote in the first column
% listing the affiliations and the copyright notice.
% The command takes one argument, which is text to display at the start of the footnote.
% The \mlsysEqualContribution command is standard text for equal contribution.
% Remove it (just {}) if you do not need this facility.

\printAffiliationsAndNotice{}  % leave blank if no need to mention equal contribution
% \printAffiliationsAndNotice{\mlsysEqualContribution} % otherwise use the standard text.

\section{Introduction}

% Large language models (LLMs) have achieved remarkable performance in text generation tasks \cite{brown2020language, achiam2023gpt, touvron2023llama}, where each input prompt undergoes a prefill phase to encode the context, followed by a decoding stage to autoregressively generate a new token step by step.

Large language models (LLMs) are extensively used in generative tasks, including chatting (e.g., ChatGPT~\cite{chatgpt}, Deepseek~\cite{deepseek2025}, Claude~\cite{claudeai2025}), code generation (e.g., GitHub Copilot~\cite{github_copilot2025}, Trae\cite{traeai2025}, Cursor~\cite{cursor2025}).
Each input prompt first undergoes a prefill phase to encode the context and generate the initial output token, followed by a decoding stage that autoregressively generates subsequent tokens step by step. 

Nevertheless, there are also many \textbf{prefill-only applications} of LLMs~\cite{du2025prefillonly}, such as classification~\cite{wang2018glue, gholamian2024llm, vajjala2025text}, question answering~\cite{talmor-etal-2019-commonsenseqa, hendrycks2020measuring, phan2025humanity}, recommendation~\cite{wang2023enhancing, wu2024survey, firooz2025360brew}, and data labeling~\cite{he2023annollm, lan2024depression, zhang2023llmaaa}. These workloads use only the embeddings from the final layer, either to produce a single token or as input for downstream tasks. This process does not involve the decoding stage or require generating multiple tokens, so the KV cache is not reused for extended decoding. Therefore, storing the KV cache is unnecessary, and only the prefill stage of LLM inference needs to be executed.
For example, in a question answering application, the input prompt could be \textit{``What is the capital of France? A. Berlin, B. London, C. Paris, D. Rome. Your answer is:"}, and the LLM only needs to generate a single answer token (i.e., \textit{A, B, C, or D}).

Furthermore, LLMs can also serve as text encoders to extract general-purpose sentence embeddings, excelling in text representation tasks \cite{lee2024gecko, behnamghader2024llm2vec, lee2024nv, li2024making}. In such tasks, only the prefill stage is used to encode the input sentences, with either the hidden states of the last token \cite{wang2023improving, lei2024meta} or the pooling of all token representations~\cite{li2024your,lei2025enhancing} as the sentence embedding, without involving decoding stage of LLM inference.

Central to the prefill stage of LLM inference is the self-attention mechanism, which enables LLMs to capture dependencies and relationships across different positions within a sequence. Attention maps, computed as the product of Query (Q) and the transpose of Key (K), encode the relevance of each position to others. However, the quadratic time complexity of this computation with respect to sequence length poses a significant performance bottleneck. 

%Various approaches have been proposed to accelerate self-attention inference by reducing computation. Most prior efforts~\cite{ham20203, wang2021spatten} exclude less important tokens from the inputs. Some efforts~\cite{ying2021lazyformer, xiao2019sharing, bhojanapalli2021leveraging} share attention maps calculated in prior layers in multiple subsequent layers. %or reuse the attention maps by identifying semantically similar inputs \cite{feng2023\name}. 
%Yet, these efforts achieve shorter time at the cost of significant loss in model accuracy, especially in complex tasks that require full-contextual information. Some efforts \cite{lu2022dynamic, kitaev2020reformer} reduce computation cost by introducing sparsity into the self-attention. Some efforts \cite{he2024matters, men2024shortgpt} remove the attention layer or transformer layer to accelerate LLM inference. However, they either change model architecture or require specialized hardware to achieve expected speedup. 

\begin{figure*}[t]
    \centering
    \label{fig:example}
    \noindent
    \begin{minipage}[b]{0.24\textwidth}
        \includegraphics[width=\textwidth]{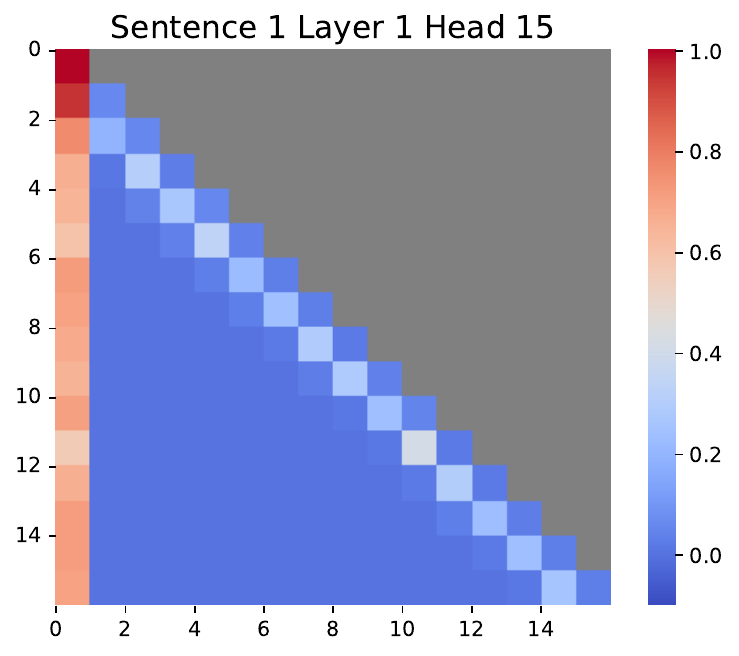}    
        % \caption{figures}
    \end{minipage}
    \begin{minipage}[b]{0.24\textwidth}
        \includegraphics[width=\textwidth]{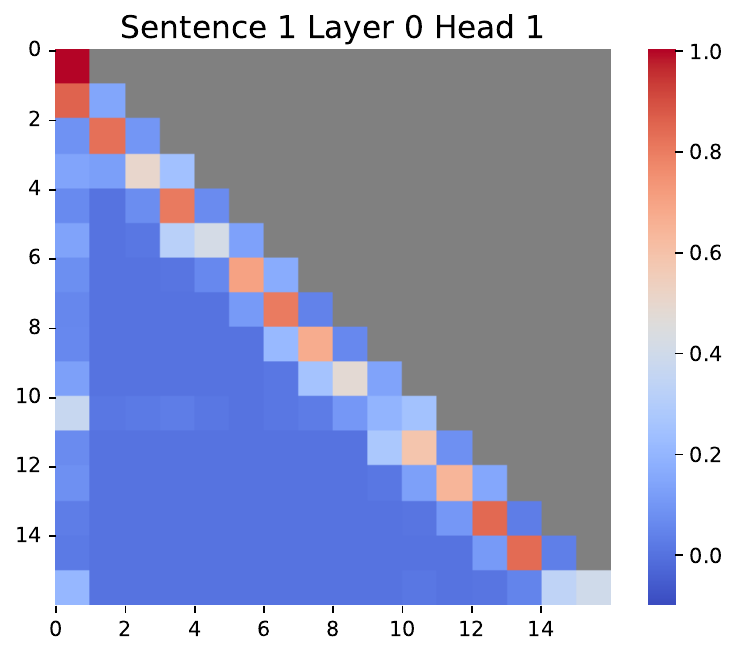}
        % \caption{figures}
    \end{minipage}
    \begin{minipage}[b]{0.24\textwidth}
        \includegraphics[width=\textwidth]{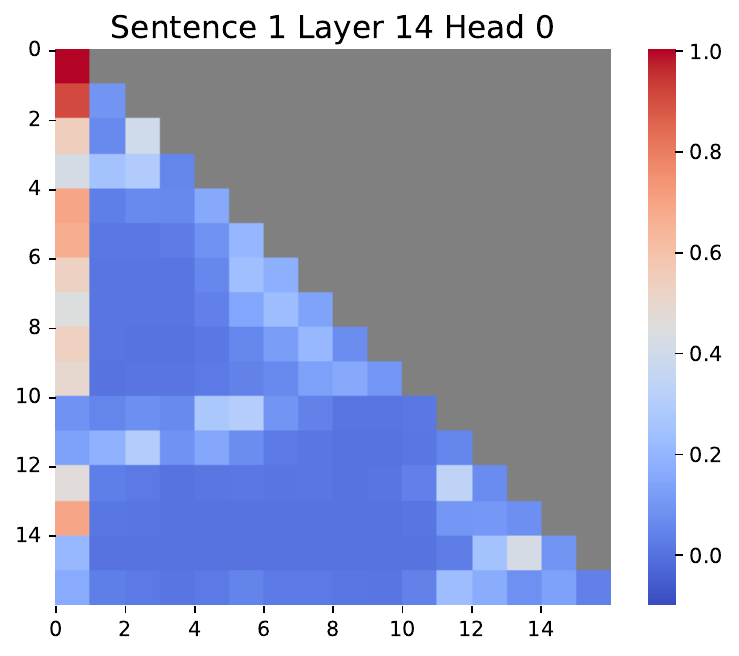}
        % \caption{figures}
    \end{minipage}
    \begin{minipage}[b]{0.24\textwidth}
        \includegraphics[width=\textwidth]{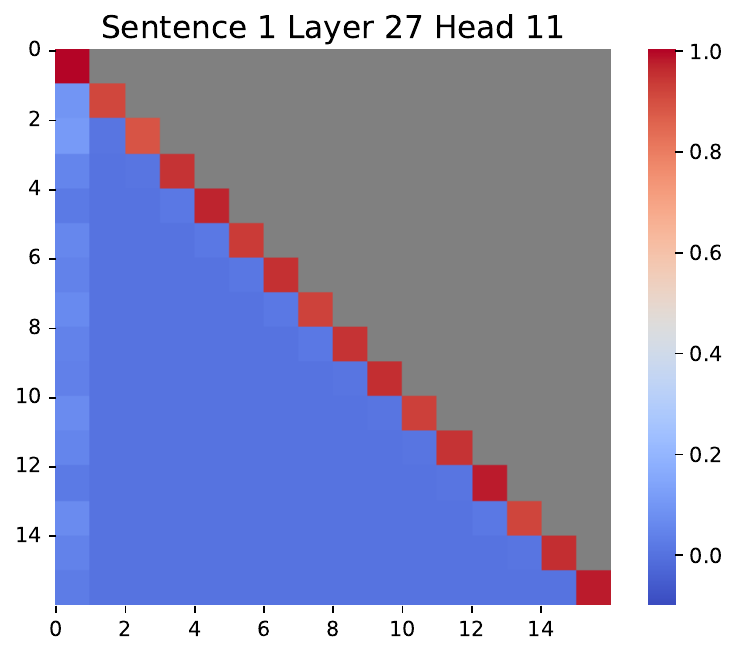}    
        % \caption{figures}
    \end{minipage}

    \begin{minipage}[b]{0.24\textwidth}
    \includegraphics[width=\textwidth]{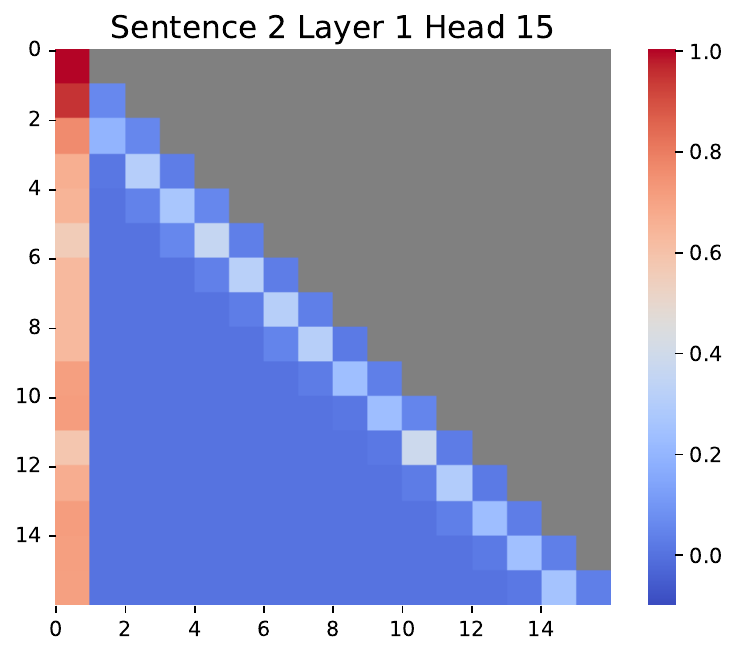}    
    % \caption{figures}
    \end{minipage}
    \begin{minipage}[b]{0.24\textwidth}
        \includegraphics[width=\textwidth]{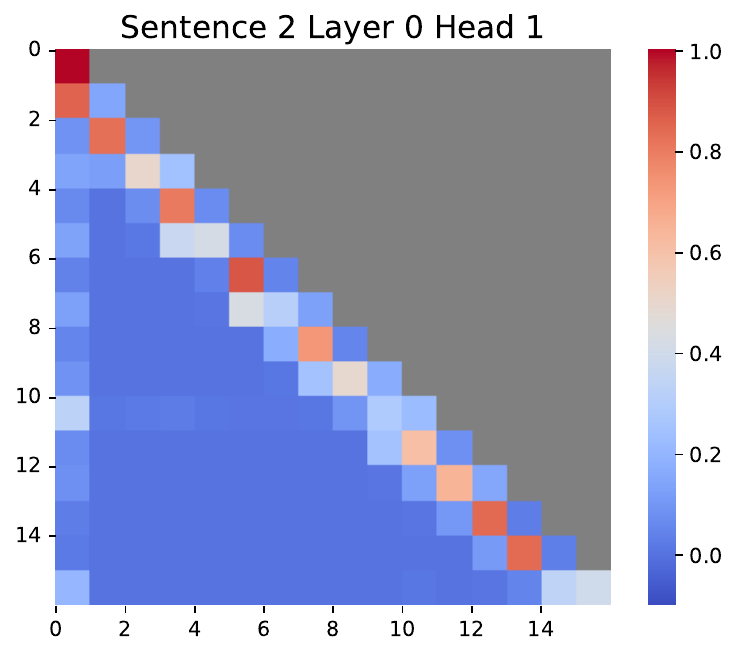}
        % \caption{figures}
    \end{minipage}
    \begin{minipage}[b]{0.24\textwidth}
        \includegraphics[width=\textwidth]{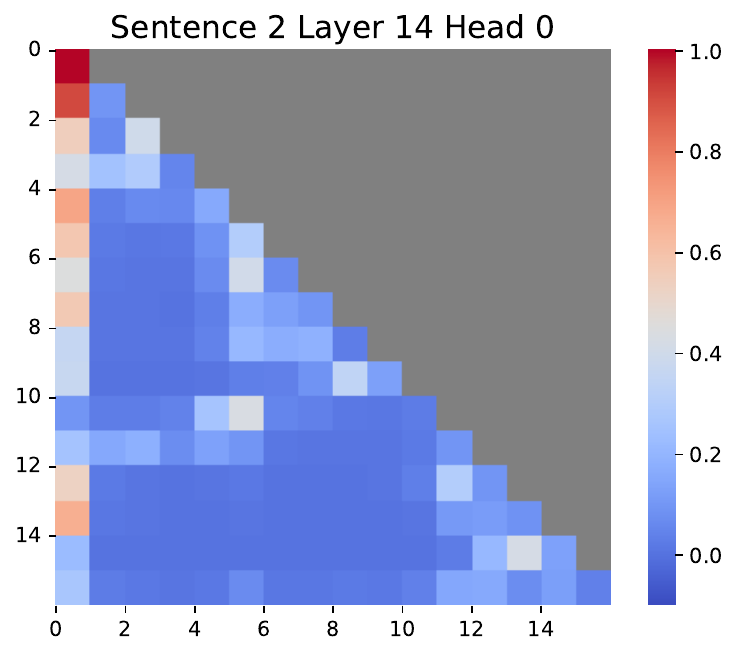}
        % \caption{figures}
    \end{minipage}
    \begin{minipage}[b]{0.24\textwidth}
        \includegraphics[width=\textwidth]{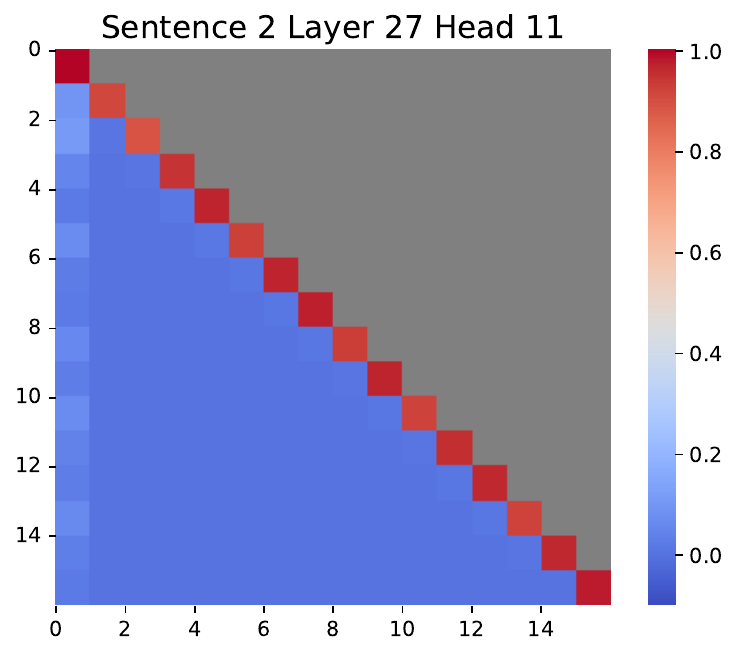}    
        % \caption{figures}
    \end{minipage}

    \caption{
    Visualization of the attention maps in Llama-3.2-3B over two sentences, each with a length of 32. 
    Sentence 1 is \textit{``This sentence: `you should never do it.' means in one word:''}. Sentence 2 is \textit{``This sentence: `how do you do that?' means in one word:''}. The plots
reveals that although Sentence 1 and Sentence 2 have different meanings, their attention maps at different layers and different heads are similar.
}
\label{fig:example}
\end{figure*}

%%%In this paper, we find that semantically different input sentences can have high similarity in their attention maps at different layers or different heads during the inference computation. By pre-storing (or caching) these similar attention maps into a database utilizing the emerging big memory system (called \textit{attention map database}), we can save self attention computation using the attention cache to retrieve similar attention maps. Our method includes two main steps: search engine and online inference. First, we generate the feature vector of the input sentence, which is embedded by a neural network. We call it \textit{Feature Projector}. This feature vector is used to retrieve the index which associated attention maps have the highest similarity to the input sentence. Second, we use the index to prefetch the attention maps of all layers from the attention map database. Those fetched attention maps are reused in the computation of self attention during inference.

In this paper, we observe that semantically different input sentences can exhibit high similarity in their attention maps across layers or heads during inference. By pre-storing (or caching) these similar attention maps in a vector database (called attention map database) or other memory systems such as SRAM, DRAM or HBM, we can retrieve and reuse them to reduce self-attention computation. For example, consider the two sentences shown in Figure~\ref{fig:example}. Sentence 1 is \textit{``This sentence: `you should never do it.' means in one word:''}. Sentence 2 is \textit{``This sentence: `how do you do that?' means in one word:''}. Although the two sentences have different semantics, their attention maps at different layers and different heads are similar, which indicates they have similar relevance at each token position. 
Consequently, the attention maps computed for Sentence 1 can be reused for Sentence 2. 
Building on this interesting insight, we introduce \textit{\name}, a framework designed to accelerate self-attention computation. Given that reusing attention maps eliminates the need for storing and computing key and query states in the KV cache, \name primarily focuses on accelerating the prefill stage of LLM inference, rather than the decoding stage, which necessitates the storage of past key and value states.

Implementing \name presents two key challenges. 
The first challenge lies in finding an effective data representation. Both the representations of input sentences and attention maps in LLMs are high-dimensional tensors. Therefore, it is practically infeasible to find similar attention maps by directly comparing the representations of input sentences. 
%To identify similar attention maps, we design a proper data representation by embedding network. The embedding network must be lightweight such that its overhead plus the search in the  attention map database (a key-value store) is smaller than the cost of self-attention computation. 
Instead, we design a lightweight embedding neural network to represent attention maps efficiently. This network must be computationally inexpensive so that its overhead, combined with the search in the attention map database, remains lower than the cost of self-attention computation.

The second challenge is the high cost of memory accesses when storing and fetching pre-populated attention maps.  
A large attention map database improves search hit rates but leads to sparse memory accesses, as accesses to attention maps exhibit poor spatial and temporal locality. 
Additionally, modern deep learning frameworks like PyTorch require tensors to be placed in consecutive memory addresses to enable vectorized data accesses for Single Instruction Multiple Data (SIMD) operations. Therefore, once a tensor is fetched from the pre-populated database, it must be copied to a consecutive memory buffer before being loaded to the processor, incurring two memory reads and one write per fetch. To reduce memory access overhead, we first store all attention maps of a layer as a single file object, and arrange the attention maps of neighboring layers continuously in the database, enhancing spatial and temporal locality. Then, \name eliminates expensive tensor copying through memory mapping between a consecutive virtual-memory space and scattered physical addresses of individual tensors.

When the memory footprint exceeds GPU capacity, we leverage CPU to demonstrate our approach in \name. Generating LLM embeddings for large-scale text corpora often surpasses GPU memory limits. In applications such as recommendation systems, which involve processing billions of text chunks, throughput across many concurrent instances is more critical than per-instance latency. For such workloads, CPUs offer better efficiency in time, energy and cost. Therefore, it is meaningful and valuable to evaluate the performance of AttnCache on CPUs.
Our evaluation shows that AttnCache achieves an average of 1.2× end-to-end and 2× attention speedup on CPU, and 1.6× end-to-end and 3× attention speedup on GPU, with negligible accuracy loss.

%We use CPU for evaluation, because some scenarios make LLMsembeddings of large text corpus, and GPU cannot bring enough memory capacity. For example, the recommendation/RAG systems get representations of billions of text pieces. Such scenarios do not need to get LLMsembeddings on a single instance fast, but must compute on a lot of instances fast. In these scenarios, using CPU machines to compute is both time and energy efficient than GPU machines. We use CPU machines to demonstrate our idea in \name. Extensive experiments show that \name with transformer-based LLMs, such as Llama-3-8B, Llama-2-7B, and Mistral-7B, on seven Semantic Textual Similarity (STS) tasks enables 1.2$\times$ speedup on average with  2\% loss in the Spearman correlation score.  %(2) If the attention maps of two sentences are very similar at a certain layer, they are also similar in other layers of the LLM. (3) The length of the token sequence plays an important role in attention maps reuse.

%We evaluate \name on CPUs, as generating LLMsembeddings for large text corpora often exceeds GPU memory capacity. In scenarios like recommendation and retrieval-augmented generation (RAG) systems, which process billions of text pieces, throughput across many instances is more critical than single-instance speed. CPUs offer greater efficiency in both time and energy for such workloads.

%%%%%%%%%%%%%%%%%%%%%%%%%%%%%%%%

\begin{figure*}[t]
  \centering
  \includegraphics[width=1.0\linewidth]{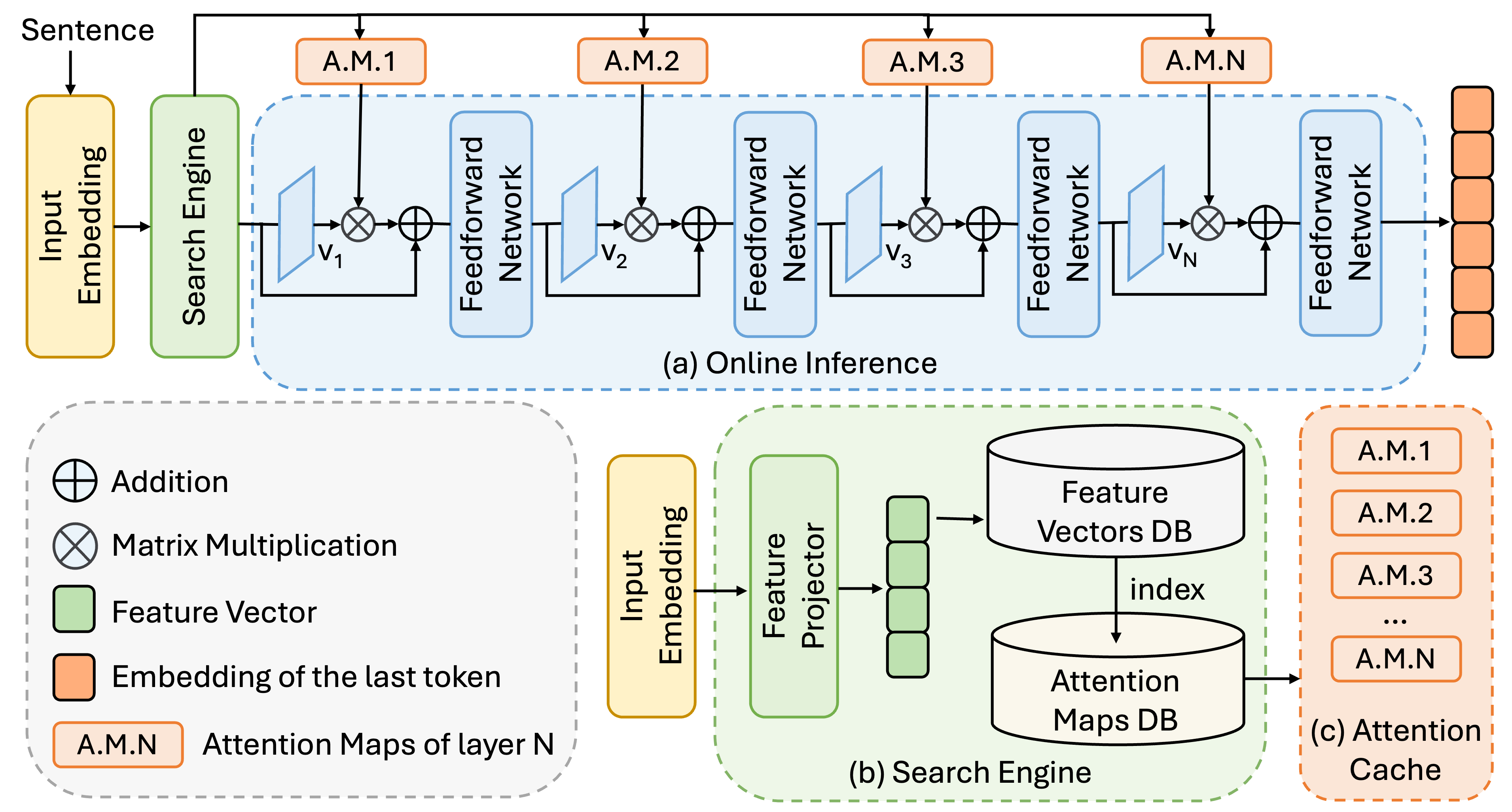}
  \caption{\name overview. The search engine will identify the index of the sentence that produces the most similar attention maps based on the feature vector of the current input sentence and prefetch attention maps for each layer from the attention map database using the index. These fetched attention maps are stored in the attention cache and reused for the matrix multiplication calculation with value projection during the self-attention computation.}
  \label{fig:attncache}
\end{figure*}
\vspace{0mm}

\section{Related Work}

\paragraph{Sentence Embedding.} Sentence embeddings encodes the semantic information of sentences into high-dimensional vector representations. Prior works~\cite{li2024your,muennighoff2024generative, ni2021sentence} have demonstrated the capability of LLMs to generate high-quality sentence embeddings. Recent studies~\cite{zhuang2024promptreps, qin2023large, zhang2024simple} have explored converting LLMs into sentence encoders without additional training. To enhance embedding quality, prompt-based techniques have gained traction. MetaEOL~\cite{lei2024meta} uses multitask prompts to generate general-purpose embeddings. The research by~\cite{jiang2023scaling} illustrates how to extract a sentence embedding by prompting  LLMs with the instruction \textit{``This sentence: `[text]' means in one word:''}. In this work, we leverage LLMs to generate sentence embeddings without fine-tuning.

\paragraph{LLM inference acceleration.} 

Most KV cache optimization approaches~\cite{beltagy2020longformer, zhang2023h2o, oren2024transformers} focus on accelerating the LLM decoding phase by reducing redundancy in Key and Value matrices. StreamingLLM~\cite{xiao2023efficient} identifies ``attention sinks'' and keeps initial and recent tokens' KV to anchor attention computation, while FastGen~\cite{ge2023model} prunes tokens during decoding by profiling attention heads. However, these approaches do not reduce the prefill costs. In contrast, several recent efforts~\cite{wu2024retrieval, jiang2024minference, tang2024razorattention} center on optimizing the LLM prefill phase, benefiting tasks like sentence embedding generation.
PromptCache ~\cite{gim2024prompt}  and ChunkAttention \cite{ye2024chunkattention} reduce time-to-first-token latency by sharing KV tensors of common prompt prefixes. Other acceleration approaches~\cite{he2024matters, men2024shortgpt, song2024sleb, zhang2024finercut} focus on removing the redundant attention or transformer layer in LLMs. Token pruning~\cite{ham20203, wang2021spatten} reduces computation by excluding less important tokens from the input, while layer-wise reuse~\cite{ying2021lazyformer, xiao2019sharing, bhojanapalli2021leveraging} reduces computation by sharing attention maps calculated in prior layers in multiple subsequent layers.
These approaches complement \name and can further enhance memory efficiency.

% Various techniques have been proposed to accelerate self-attention inference through computation reduction. 

% Token pruning~\cite{ham20203, wang2021spatten} reduces computation by excluding less important tokens from the input, while layer-wise reuse~\cite{ying2021lazyformer, xiao2019sharing, bhojanapalli2021leveraging} reduces computation by sharing attention maps calculated in prior layers in multiple subsequent layers. Despite their effectiveness in accelerating self-attention inference, these techniques often cause significant loss in model accuracy, especially in complex tasks that require full-contextual information. 

\paragraph{Reuse mechanism in Neural Networks.} The reuse mechanism exploits the inherent redundancy in neural networks to enhance efficiency. 
Prior works~\cite{ning2019adaptive, ning2019deep,wu2022drew, kopuklu2019convolutional} have explored reusing similar computation results to improve performance. \citet{silfa2019neuron} accelerate RNN training by reusing neuron outputs. 
Studies~\cite{bhojanapalli2021leveraging, xiao2019sharing} have shown that transformer attention maps \cite{vaswani2017attention} exhibit similar distributions across adjacent layers. 
Many prior efforts~\cite{hunter2023fast, xiao2019sharing, bhojanapalli2021leveraging, ying2021lazyformer, liao2024beyond} focus on sharing computed attention weights across multiple layers for the same input sequence. However, this approach may introduce dissimilar attention maps, which can degrade performance. In contrast, our work efficiently reuses similar attention maps across different sequences, overcoming the limitations of intra-sequence reuse.

%Many prior efforts \cite{hunter2023fast, xiao2019sharing, bhojanapalli2021leveraging, ying2021lazyformer, liao2024beyond} focus on sharing computed attention weights across multiple layers for the same input sequence. However, this can introduce discrepancies in attention maps, degrading performance.

%Recent effort \cite{ feng2023\name} exploits the similarity between different sequences by computing the similarity of hidden states at each layer to search for similar attention maps. 

\section{Methodology}

%\name consists of two major components: the search engine and online inference. 
As shown in Figure~\ref{fig:attncache}, given an input sentence, \name embeds it into a feature vector using a lightweight neural network (feature projector). The feature vector is used to retrieve the index of the attention maps that have the highest similarity to the input sentence. Then, the search engine uses the index to fetch the corresponding attention maps from the attention map database. The fetched attention maps are used in the self-attention computation during online inference, while the prefill stage in LLM inference is utilized to generate the sentence embedding.

% We use the prefill (initial processing of the input sequence)  stage of LLM inference to genetate the sentence embedding.  

% \textcolor{red}{(pending...)}

% \textcolor{red}{(pending to mention that we focus on prefill.)}

%The search engine includes feature projector and feature vector database of input embedding. By utilizing the feature vector of the input embedding, the search engine retrieves the index of similar attention maps from the feature vector database and prefetches the corresponding attention maps from the attention map database using this index. These prefetched attention maps will be used in the self attention computation during online inference. We discuss details as follows. \textcolor{red}{(mention that we focus on prefill?)}

%First, we generate the feature vector of the input sentence, which is embedded by a neural network. We call it \textit{Feature Projector}. This feature vector is used to retrieve the index which associated attention maps have the highest similarity to the input sentence. Second, we use the index to prefetch the attention maps of all layers from the attention map database. Those fetched attention maps are reused in the computation of self attention during inference.

\subsection{Search Engine}

As illustrated in Figure \ref{fig:example}, two sentences with completely different semantics can produce highly similar attention maps.
Because input sentences are represented as high-dimensional hidden states, directly comparing these representations provides little insight into the similarity of their corresponding attention patterns.
To overcome this limitation, \name uses the feature vector of input hidden states, which is embedded by the feature projector. By searching for similar feature vectors, \name can efficiently retrieve input embeddings that yield similar attention maps.

\begin{algorithm}[tb]
\caption{Search Engine}
\begin{algorithmic}[1]  
\STATE \textbf{Input:} Sentence $S$, Threshold $\theta$;
\STATE \textbf{Output:} Attention Cache $attn\_cache$, \\ \hspace*{3.5em} Input embedding $h$;
\STATE \textbf{Function} \textsc{search\_engine}($S$, $\theta$)
  \begin{ALC@g}  
  \STATE $h \gets \text{encode}(S)$
  \STATE $f \gets \text{feature\_projector}(h)$
  \STATE $(idx, sims) \gets \text{VecDB.search}(f)$
  \STATE $attn\_cache \gets []$
  \IF{$sims \geq \theta$}
      \STATE $n \gets num\_layers$
      \STATE $ams \gets \text{AttnMapsDB.get}(idx, n)$
      \STATE $attn\_cache.\text{append}(ams)$
  \ENDIF
  % \STATE \textbf{return} $(attn\_cache, h)$
  \end{ALC@g}
\STATE \textbf{return} $(attn\_cache, h)$
\end{algorithmic}
\end{algorithm}

%We share the weights of feature projector \textbf{W} and

%%%%After finishing training, the feature vectors are stored in \textcolor{red}{the database} so that we can find the similar attention maps based on the index of similar feature vector.

\paragraph{Feature Projector.}

% To quantify the similarity of input embeddings, we use a feature projector, which is an embedding network. Two input embeddings are matched during a search if their feature vectors are similar (in terms of the similarity score defined later). 

% The feature vector is essentially an internal representation of the input embedding to capture similarity, and the feature projector learns this representation through training such that the input embeddings with similar attention maps have similar feature vectors. By searching for similar feature vectors, we are able to find input embeddings producing similar attention maps. Besides, the feature projector allows us to map input embeddings into a lower-dimensional representation, thus reducing the search space and computation complexity of measuring the similarity.

\begin{figure}[t] 
    \centering
    \includegraphics[width=0.9\linewidth]{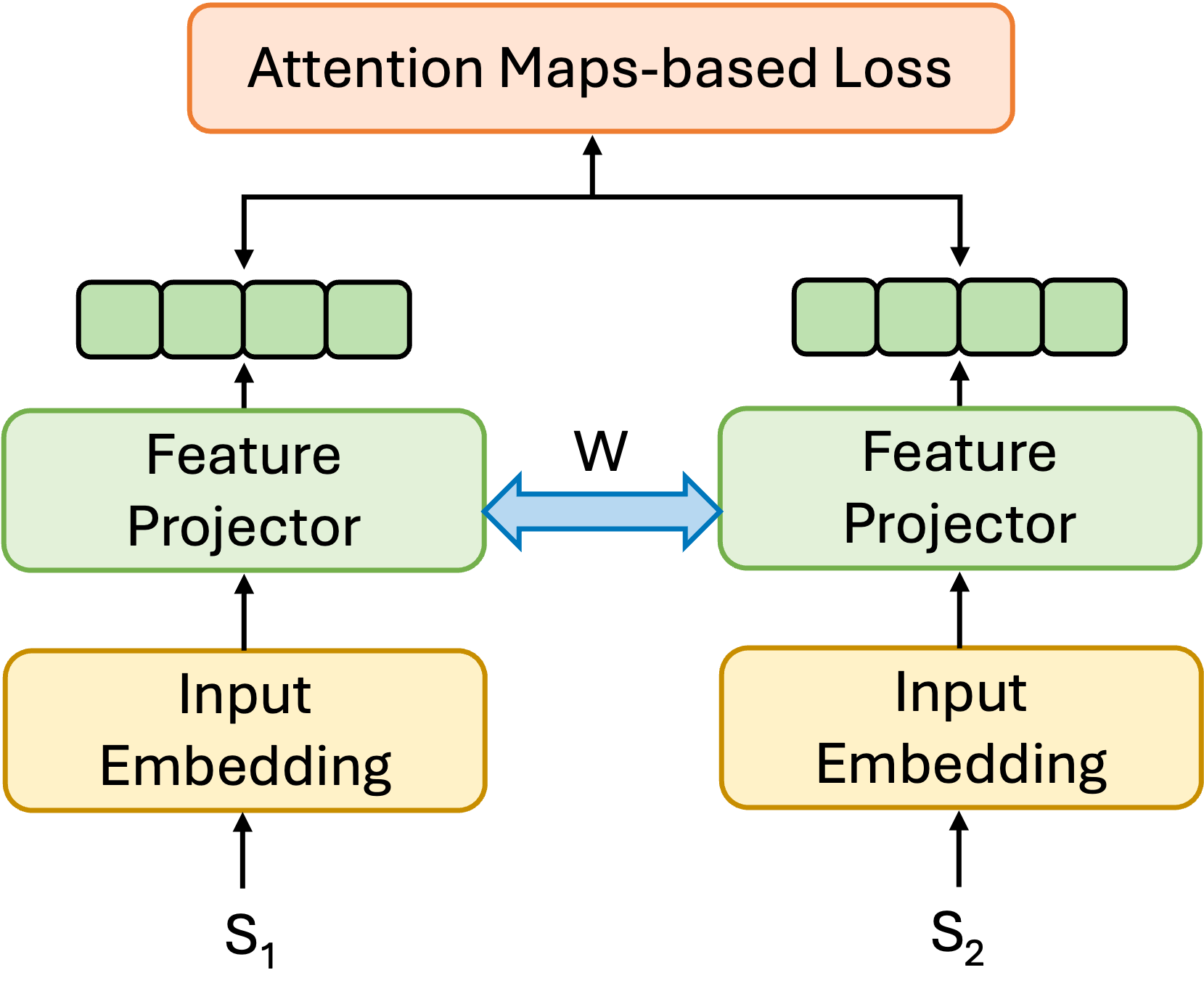} 
    \caption{The training of the feature projector. The feature projector maps input embedding of a sentence \textbf{S} into a feature vector. Then we train the feature projector using the attention maps-based loss function.} 
    \label{fig:feature_projector}
    \vspace{-10pt}
\end{figure}

%The network structure of Feature Projector is important to the accuracy and efficiency of the search process. 
We use two layers of Multi-Layer Perceptron (MLP) as the feature projector, which maps the input embedding to a feature vector with lower dimension size. The network structure of Feature Projector is important to the accuracy and efficiency of the search process. Compared with other embedding models, such as convolutional neural network or transformer, MLP is lightweight with less computational complexity and shorter inference time. Training the feature projector is challenging due to a lack of labeled data. Deciding the similarity between input embeddings and labeling them as similar or not is prohibitively expensive. 
%Following \cite{feng2023\name}, 
We use the Siamese network \cite{koch2015siamese}, which contains two identical feature projectors and shares the same weights, as shown in Figure \ref{fig:feature_projector}.  

% Once the Siamese network finishes training, it is used as a feature projector. The Siamese network is trained to minimize the distance between feature vectors  whose attention maps have high similarity.

During each training iteration, two input embeddings are used as input to the two identical feature projectors in the Siamese network. After getting the feature vectors, the Euclidean distance (i.e. L2-norm) is calculated as follows. 
\begin{equation}
\small
\label{eq:metric}
    \hat{y} = ||f_{\textbf{W}}(\textbf{X}_{1}) - f_{\textbf{W}}(\textbf{X}_{2}))||_{2}
\end{equation}
where \textbf{X} is the input embedding, %\textbf{w} is the weights of feature projector; 
$f_{\textbf{W}}$ is the feature projector, and $||.||_{2}$ is the L2 norm. Besides, we measure the similarity score using the attention maps and the sequence length of tokens, which associate with the two input embeddings. We use the metric as the labels for training the feature projector based on the average distance of heads, which is defined as follows. 
\begin{equation}
\small
\label{eq:labels_for_feature_projector}
    y = \frac{1}{n} \times \alpha \sum_{p=1}^n\frac{1}{2}|| \textbf{A}_{1}[p,:] - \textbf{A}_{2}[p,:] ||_{2}  + ||s_{1} - s_{2}||_{1}
\end{equation}
where \textbf{A} denotes the attention map, $n$ indicates the number of head, \textbf{A}$[p,:]$ is the $p^{th}$ row of the attention map, $||.||_{1}$ is the L1 norm, $s$ denotes the length of input token sequence, and $\alpha$ is the hyperparameter to control the relative importance of the similarity of the attention maps and the token length. In addition to the inherent similarity of the attention maps, the token sequence also plays an important role in determining whether two attention maps are similar. When the token sequences of two attention maps are very different in length, even if the attention maps are similar, they cannot be used directly in \name,  otherwise it may cause a large inference error. The final loss function of the feature projector is defined as follows.

\begin{equation}
\small
L = 
\begin{cases} 
0.5(\hat{y} - y)^2 & \text{if } |\hat{y} - y| < 1 \\
|\hat{y} - y| - 0.5 & \text{if } |\hat{y} - y| \geq 1
\end{cases}
\end{equation}
We use Smooth L1 Loss~\cite{girshick2015fast} as the loss function, which is able to balance the effects of outliers. The training process iteratively updates the parameters of the feature projector to minimize the loss function. 
% Using the training process described above eliminate the need for labeling input embeddings and attention maps manually.

%%%%%%%%%%%%%%%%%%%%%%%%%%%%%%%s

\begin{figure}[t]
        \centering
        \includegraphics[width=\linewidth]{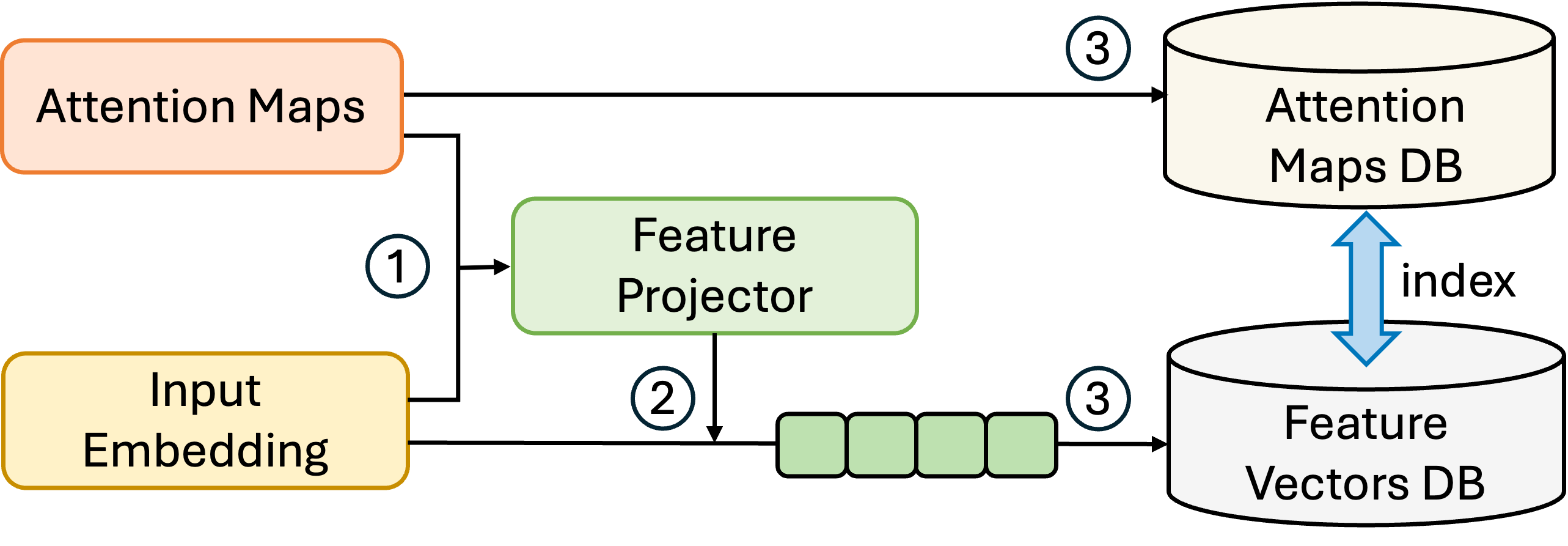}
	\caption{Databases building include three steps. 1. Train the feature projector with input embeddings and attention maps; 2. Embed the input embeddings to feature vectors; 3. Store the feature vectors and attention maps to their respective databases. Both databases share the same index.}
	\label{fig:database_building}
\end{figure}

\paragraph{Databases.} 
%After completing the training of the feature projector, the feature vector database and attention map database must be built.
To minimize the costly search for attention maps, we construct an indexed database, where feature vectors are stored and indexed for fast search. In essence, the feature vector database is a key-value store where the key and value are the feature vector and its index. Figure~\ref{fig:database_building} illustrates the process of building the databases. The attention maps associated with the feature vectors are stored in the attention map database, where the key is the index and value is the attention map. Both databases have the same index.

%%%%%%%%%%%%%%%%%%%%%%%%%%%%%%%%%%%%%%%%%%%%

%The functioin \texttt{search\_engine} in Algorithm \ref{alg:search_engine} shows the pipeline of fetching attention maps from the attention map database. The input sentence is embedded by input embedding (Line 2). Input embeddings includes tokenization of the sentences, position encoding, layer normalization. Then the result is mapped into a feature vector with lower dimension(Line 3). The feature vector are used for querying the VecDB, which stores the feature vector of the pre-collected input embedding mapped by the feature projector. After the query completes, the indices that have closest similarities with the feature vectors in the VecDB are returned(Line 4). The similarity between the feature vector in VecDB and the query feature vector is measured using Euclidean distance. If the similarity is larger than or equal to a given threshold $\theta$, the corresponding index $idx$ is used to fetch the attention maps from the attention map database. 

% Algorithm \ref{alg:search_engine} depicts the search engine. 
Algorithm 1 illustrates the process of finding the most similar attention maps, referred to as \texttt{search\_engine}. The input sentence is embedded by input embedding (Line 2). The input embedding includes tokenization of the sentence, position encoding, and layer normalization. Then the result is mapped into a feature vector with lower dimension (Line 3). The feature vector is used for querying in the feature vector database. After the query, the indices that have the highest similarity to the feature vector are returned (Line 4). When the similarity is not less than the threshold $\theta$, the corresponding index $idx$ is used to fetch attention maps from the attention map database. 

The retrieved index $idx$ corresponds to a sentence $S$ whose attention maps are similar to those of the current input sentence.
The corresponding attention maps are fetched for all layers of the LLM and stored in a contiguous memory region, referred to as the attention cache.
Specifically, these attention maps are used in the matrix multiplication calculation with value projection in \texttt{online\_inference}. All layers of the attention maps are fetched for the computation of self-attention before the LLM inference starts.  %%%Instead of computing the similarity and fetching the attention maps for each layer as in \cite{feng2023\name}, \name retrieves the attention maps only once before online inference.

\subsection{Online Inference}

Algorithm \ref{alg:online_inference} illustrates online inference with \name. %A typical transformer layer contains two heavy operations blocks: self attention calculation (Line 3-16) and feed forward transformation (Line 17).  Our proposed method is used to accelerate the self attention block, and the feed forward block will be kept as normal computation.
In the attention block of each layer, the value projection is computed. If similar attention maps are found, the attention output can be obtained by multiplying the attention maps by the value $v$. Thus, finding similar attention maps and reusing them in the self-attention calculation leads to performance benefits.  

However, \name cannot always find similar attention maps. For those hidden states with low similarities, the attention maps must be calculated at each layer during the inference, which means the query, key, rotary positional encoding, and softmax normalization must be computed. In this regard, \name does not bring benefit in inference speed, and instead degrades performance due to its search overhead. However, given a batch of inferences, as long as the success rate of retrieving for all inferences is high, the overall inference is still accelerated. 

%%With \name, such overhead can be offset by the performance benefit. The final performance benefit will be the sum of the above two. 

\begin{algorithm}[tb]
\caption{Online Inference}
\label{alg:online_inference}
\begin{algorithmic}[1]
\STATE \textbf{Input:} Attention Cache $attn\_cache$, \\ \hspace*{2.8em} Input embedding $h$;
\STATE \textbf{Output:} Hidden states of last layer $h$;
\STATE \textbf{Function} \textsc{online\_inference}($attn\_cache$, $h$)
  \begin{ALC@g}
  \FOR{$l$ in range($num\_layers$)}
      \STATE $residual \gets h$
      \STATE $v \gets \text{v\_projection}(h)$
      \IF{$attn\_cache$ is not \texttt{NULL}}
          \STATE $attn\_map \gets attn\_cache[l]$
          \STATE $h \gets \text{mat\_mul}(attn\_map, v)$
      \ELSE
          \STATE $q \gets \text{q\_projection}(h)$
          \STATE $k \gets \text{k\_projection}(h)$
          \STATE $(q, k) \gets \text{rotary\_pos\_emb}(q, k)$
          \STATE $attn\_map \gets \text{softmax}(q, k)$
          \STATE $h \gets \text{mat\_mul}(attn\_map, v)$
      \ENDIF
      \STATE $h \gets residual + h$
      \STATE $h \gets h + \text{feed\_forward}(h)$
  \ENDFOR
  \end{ALC@g}
\STATE \textbf{return} $h$
\end{algorithmic}
\end{algorithm}

%%In the attention block of each layer, the value projection is computed. If the $attn\_cache$ is not null, which means  similar attention maps are found, the attention output can be obtained by multiplying $attn\_cahe$ with value $v$. Thus finding a similar attention maps and reusing it in the self-attention calculation leads to performance benefits.  However, since \name cannot be always successfully applied, for those hidden states with low similarities, the attention maps needs to be calculated from scratch at each layer during the inference, which means the query, key, rotary positional embedding and softmax normalization need to be computed. In this case, there is no performance benefit and the search overhead cannot be covered, which causes performance loss. With \name, such overhead can be offset by the performance benefit. The final performance benefit will be the sum of the above two. 

\section{Applicability of AttnCache}

% AttnCache is well-suited for tasks that rely solely on the prefill stage inference of LLMs, such as recommendation, classification, clustering, and retrieval, where LLMs are primarily used for embedding extraction. In these scenarios, rapidly generating high-quality embeddings is critical. The method proposed in this paper is effective in such contexts, achieving faster inference by reusing cached attention maps while maintaining minimal degradation in embedding quality.
AttnCache is well-suited for tasks that rely solely on the prefill stage inference of LLMs. By reusing similar attention maps, AttnCache effectively reduces three time-consuming matrix multiplications at each layer, i.e.,
\begin{equation}
\resizebox{0.92\columnwidth}{!}{$
\begin{aligned}
\mathbf{Q} &= \mathbf{X} \mathbf{W}_Q, & \quad \mathbf{Q} \in \mathbb{R}^{ L \times d_k} \\
\mathbf{K} &= \mathbf{X} \mathbf{W}_K, & \quad \mathbf{K} \in \mathbb{R}^{ L \times d_k} \\
\text{AttnMaps} &= \mathrm{softmax}\left( \frac{\mathbf{Q} \mathbf{K}^\top}{\sqrt{d_k}} \right), & \quad \text{AttnMaps} \in \mathbb{R}^{L \times L}
\end{aligned}
$}
\end{equation} 
where $L$ denotes the input sentence length in tokens, 
$\mathbf{X} \!\in\! \mathbb{R}^{L \times d},\ 
\mathbf{W}_Q, \mathbf{W}_K \!\in\! \mathbb{R}^{d \times d_k} 
$. For simplicity, we omit the notation for batch size and the number of attention heads. When reusing attention maps, \textbf{Q} and \textbf{K} are neither computed nor stored. However, during LLM decoding inference, the keys stored in the KV cache are required for each step of autoregressive token generation. Therefore, although AttnCache can accelerate inference in the prefill stage, its inability to compute and store \textbf{K} makes it unsuitable for decoding scenarios. Another limitation is the mismatch in attention map dimensions between the prefill and decoding stages. In the decoding phase, since only one token is generated at a time, the corresponding attention map has a shape of $1 \times (L + t)$ rather than $L \times L$, i.e.,
\begin{equation}
\resizebox{\columnwidth}{!}{$
\begin{aligned}
\mathbf{Q}_t &= \mathbf{X}_t \mathbf{W}_Q, & \quad \mathbf{Q}_t \in \mathbb{R}^{1 \times d_k} \\
\mathbf{K}_{\leq t} &= \text{cached Keys} & \quad  \mathbf{K}_{\leq t} \in \mathbb{R}^{(L + t) \times d_k}\\
\text{AttnMaps} &= \mathrm{softmax} \left( \frac{ \mathbf{Q}_t \mathbf{K}_{\leq t}^\top }{ \sqrt{d_k} } \right), & \quad \text{AttnMaps} \in \mathbb{R}^{1 \times (L + t) } \\
\end{aligned}
$}
\end{equation}

AttnCache currently can only accelerate computation during the prefill stage, as the reused AttnMaps have a shape of $L \times L$ and therefore cannot be applied in the decoding stage. Investigating how to reuse AttnMaps with a shape of $1 \times (L + t)$ could be a promising research direction to reduce KV cache storage and the computational cost of self-attention during LLM decoding.

% Our study is driven by the recent development of memory technologies (e.g., Compute Express Link~\cite{sharma2023introduction} or persistent memory~\cite{DBLP:journals/corr/abs-1903-05714}) that
% enable big memory systems at large scales (e.g., terabyte or petabyte scale)~\cite{cxl_large,cxl_pb}. 
%  In addition, the performance of NVMe ultra-low latency (ULL) SSD \cite{electronics13010174} (e.g., Optane SSD 800p and Z-SSD) is close to that of existing big memory solutions and can provide TB-scale (or PB-scale \cite{samsung_ssd_pb}) capacity, which meets the needs of building the databases. Specifically, we can leverage increased memory capacity to store self-attention maps. 

 % Besides, \name requires preloading attention maps datasets into the big memory systems. When new reusable attention maps need to be added, the feature projector needs to be retrained. Therefore, to improve the hit rate of attention maps reuse, incremental training is essential while expanding the attention map database. 

\section{Experiments} \label{sec:exp}

\begin{figure*}[!t]
  \centering
  \includegraphics[width=0.82\textwidth]{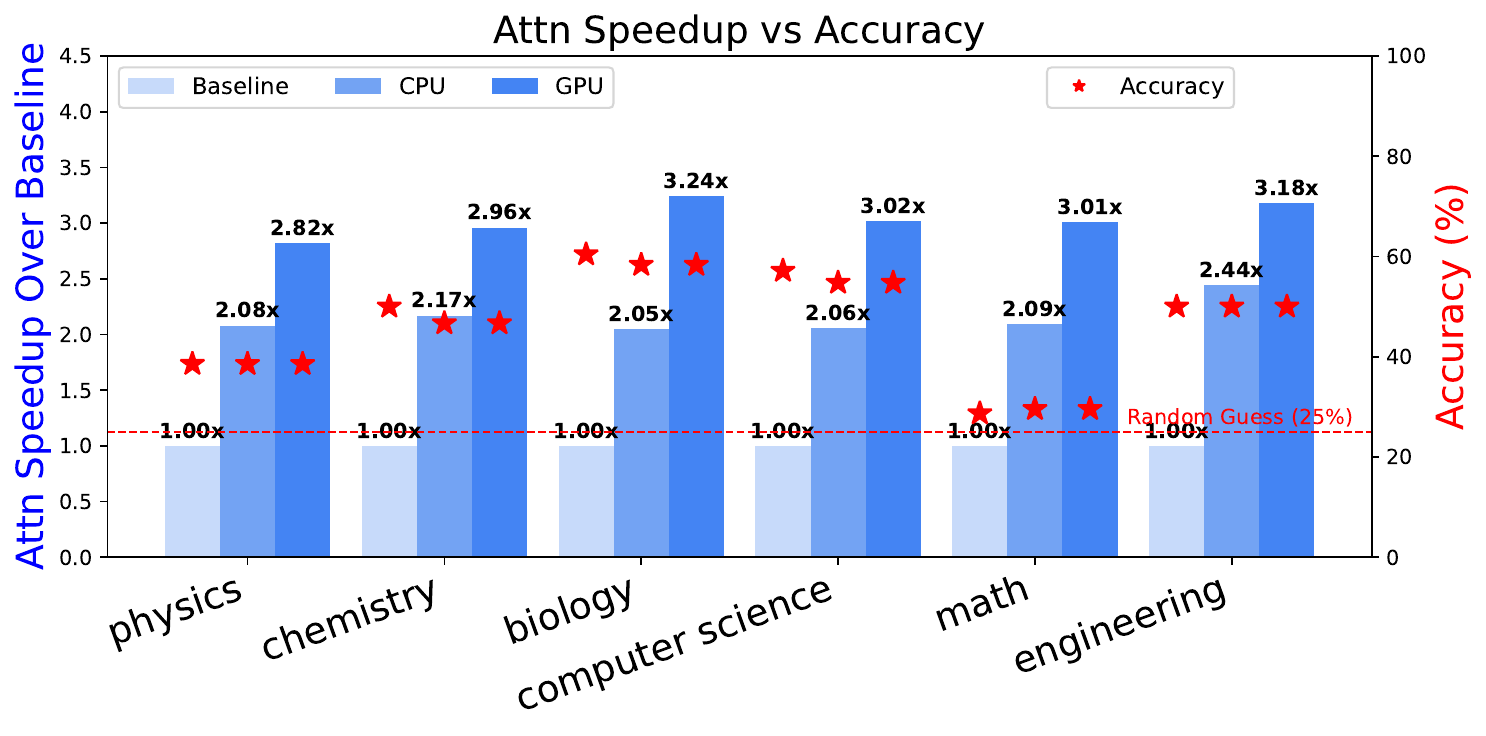}
  % \hfill
   % \vspace{0.1em} 
  \includegraphics[width=0.8\textwidth]{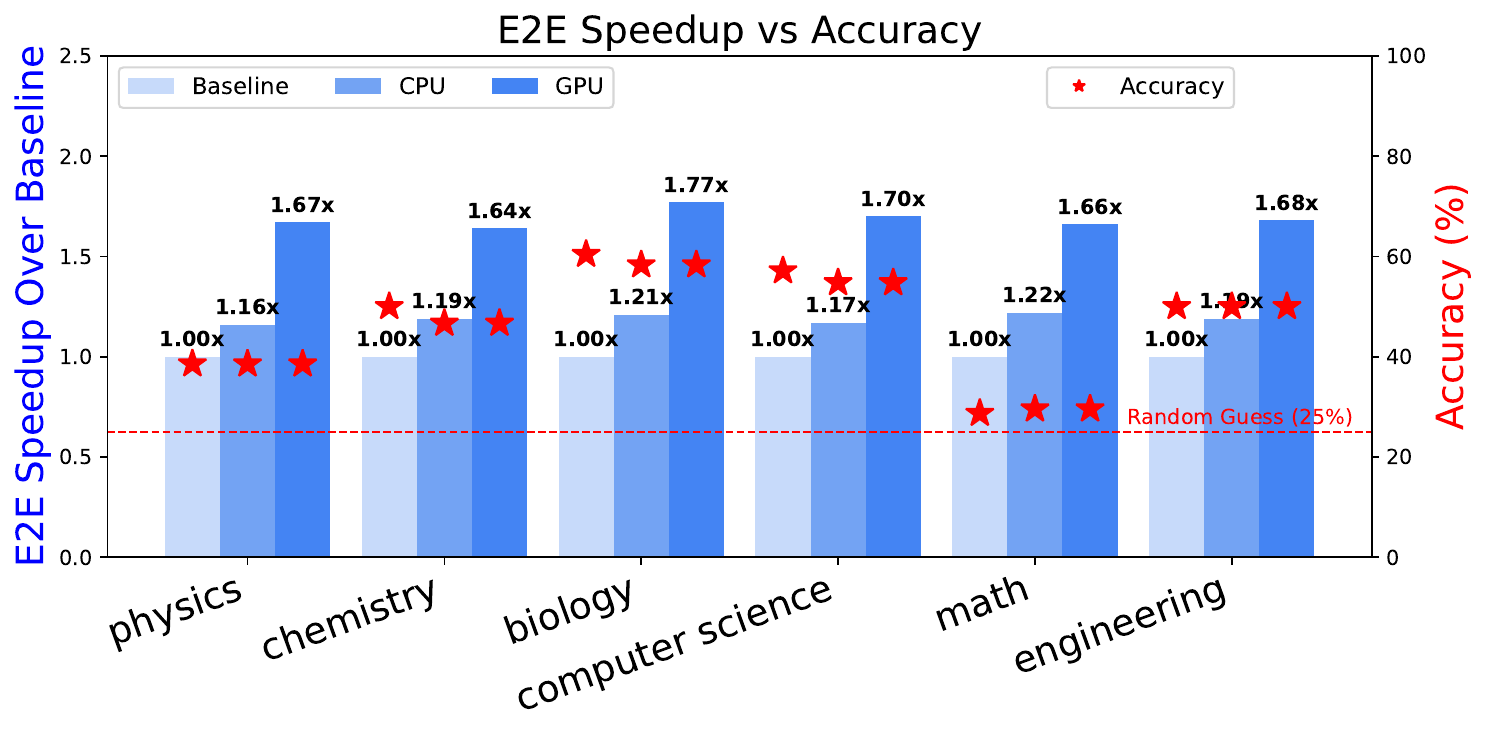}
  \vspace{-5mm}
   \caption{Attention and end-to-end speedup of Llama-3-3B with AttnCache on MMLU STEM. 
    By selecting an appropriate threshold, AttnCache achieves an average of 1.2$\times$ end-to-end and 2$\times$ attention speedup on CPU, and 1.6$\times$ end-to-end and 3$\times$ attention speedup on GPU, while incurring only a negligible drop in accuracy. }
   \label{fig:mmlu_stem_small}
\end{figure*}

\subsection{Datasets}
\label{sec:data_model}
 
We take three representative datasets, including Semantic
Textual Similarity (STS) \cite{muennighoff2022mteb} , Stanford Sentiment Treebank v2 (SST-2) \cite{wang2018glue}, and Massive Multitask Language Understanding (MMLU) \cite{hendrycks2020measuring}. STS datasets contain STS 12-16, STS-B and SICK-R \cite{lei2024meta}. The semantic similarity of each sentence pair is annotated with a score of 0-5. We use the Spearman correlation score~\cite{lei2024meta} between the ground-truth similarity scores and the predicted similarity scores as the evaluation metric. SST-2 is a binary sentiment classification dataset derived from movie reviews. Each sentence is labeled as either positive or negative. MMLU is a multi-choice dataset containing questions from 57 diverse subjects, including math, computer science, engineering, physics, and more. For SST-2 and MMLU, we use accuracy as the evaluation metric. These datasets and their outputs and labels are summarized in the Table~\ref{tab:dataset_summary}.
\begin{table}[h]
\centering
\caption{Summary of STS-B, SST-2, and MMLU with task types and labels.}
\footnotesize
\begin{tabularx}{\columnwidth}{lXXX}
\toprule
\textbf{Task} & \textbf{Type} & \textbf{Output} & \textbf{Label} \\
\midrule
STS  & Semantic similarity estimation  & Continuous score     & Similarity score (0--5)     \\
SST-2  & Sentiment classification        & Categorical label    & Positive / Negative (1 / 0) \\
MMLU   & Multitask multiple-choice QA    & Categorical label    & Multiple choice (A/B/C/D)   \\
\bottomrule
\end{tabularx}
\label{tab:dataset_summary}
\end{table}

% Examples of input sentences for each dataset can be found in Appendix \ref{input_example}.

\subsection{Models}

We conduct experiments using three representative open-source models, including Llama-2-7B \cite{touvron2023llama}, Llama-3-8B \cite{touvron2024llama3}, and Mistral-7B \cite{jiang2023mistral}. All these models are run with weights stored in full-precision (fp32) floating-point format, and evaluation is conducted using the SentEval toolkit \cite{conneau2018senteval}, measuring performance on CPU. For single-GPU scenarios, we evaluate the full-precision Llama-3-3B \cite{touvron2024llama3} and the 4-bit quantized Llama-3-8B, Deepseek-MoE-16B \cite{deepseekmoe2024} , and Qwen1.5-MoE-A2.7B \cite{qwen2024qwen}  on MMLU datasets. To assess performance under varying context lengths, we evaluate Llama-3-3B  on MMLU under different input n-shot settings. To demonstrate the generality of AttnCache, in addition to Transformer decoder-based LLMs, we also evaluate its performance on Transformer encoder-based models, such as BERT base \cite{devlin2018bert}, RoBERTa \cite{liu2019roberta}, and DeBERTa \cite{he2020deberta}. 

% Due to space limitations, the experimental results for this part can be found in Appendix~\ref{appendix:bert_perf}.

\subsection{Experimental Setting}

\begin{table*}[htbp]
\centering
\caption
% {
% Results of Different Layers Replacement on STS tasks (Spearman correlation scaled by 100x)(threshold = 0.9999)}
{
\textbf{Spearman correlation score  (in \%) across 7 STS tasks.}
% , where we replace the fixed number (e.g., 4 and 8) of attention modules in \name-f and drop the corresponding modules in AttnDrop.
}

% on Llama-2-13B and Mistral-7B. Here, Block and Attn are corresponding modules. % of the original performance are
\renewcommand{\arraystretch}{0.95}
\resizebox{\textwidth}{!}{ 
\begin{tabular}{l|ccccccc|ccc}
\toprule
\multicolumn{11}{c}{
Llama-2-7B} \\

\midrule

 \textbf{Method} &  STS12 & STS13 & STS14 & STS15 & STS16   & STS-B & SICK-R & Avg. (↑)& E2E Speedup (↑)& $\gamma$ (↓) \\
\midrule

  Full Model  &  60.88  & 73.93  & 58.30 & 70.27 & 75.46  &  73.89   &     67.44  & 68.60  & 1.00× & -- \\
  % Full Model (all samples) & 58.99 &  77.33 &  66.67&  73.55 & 73.84 &  71.72    &     69.54    &  70.23 & 1.00× & -- \\

\midrule 

SAN & 5.02 & 42.63 & 19.84& 43.49& 44.70&   18.01  &     38.71   & 30.34 & 1.45× & 0.85   \\
LazyFormer&  23.79 & 34.88 & 27.80 & 35.93 & 44.04 &  32.50  &    42.45  & 34.48 & 1.39× & 0.87  \\
% \midrule 
\name-f &  22.06 & 67.75 & 31.52 & 61.15 & 53.89 &    53.97     &     62.40    & 50.39 & 1.14× & 1.30  \\
% PromptCache &  -- & -- & -- & -- & -- & -- & --  & -- & -- &--  \\
% \midrule
\textbf{\name} &  60.59 &  73.46 &  57.97 &  69.01 &  75.38&     72.02    &    65.85 &  67.75 & 1.19× & 0.04  \\

\midrule

\multicolumn{11}{c}{
Llama-3-8B } \\

\midrule

 \textbf{Method} &  STS12 & STS13 & STS14 & STS15 & STS16 & STS-B & SICK-R   & Avg. (↑)& E2E Speedup (↑)& $\gamma$ (↓) \\
\midrule

  Full Model  &  61.57  & 76.41  &  63.23 &  75.27  &  80.41  &  75.84   & 70.45   &  71.88  & 1.00× & -- \\

  % Full Model(all samples)  & 60.68 & 80.28 & 69.54 & 76.80 & 78.10 &    74.10     &      69.44      & 72.71 & 1.00× & -- \\

\midrule 

SAN &  27.61  & 53.81  & 37.18 & 57.20  & 57.43  &   39.46      &    54.98      & 46.81 & 1.49× & 0.51  \\
LazyFormer&  27.25 & 60.37  & 36.21 & 53.85 & 59.21  & 40.30 & 48.24  & 46.49 & 1.42× & 0.60  \\
% \midrule 
\name-f &  24.89 & 51.15  & 36.19 & 67.81  & 61.39 & 48.05 & 63.77   & 50.46 & 1.16× & 1.34  \\
% PromptCache &  -- & -- & -- & -- & -- & -- & --  & -- & -- &--  \\
% \midrule
\textbf{\name} &  60.82 & 72.49  & 60.59 & 74.67 & 79.52 & 72.61 & 66.68   & 69.63 & 1.21× & 0.11  \\

\midrule 

\multicolumn{11}{c}{
Mistral-7B } \\

\midrule

 \textbf{Method} &  STS12 & STS13 & STS14 & STS15 & STS16 & STS-B & SICK-R   & Avg. (↑)& E2E Speedup (↑)& $\gamma$ (↓) \\
\midrule

 Full Model  &  63.28  & 74.89  &  61.57  &  75.64  &  81.89    & 78.26  &  69.39  &  72.13  & 1.00× & -- \\
% Full Model(all samples)  & 63.11 & 78.49 & 69.38 & 77.92 & 79.15 &    75.73  &  69.51   & 73.33  & 1.00× & -- \\
\midrule 

SAN &  25.04 & 54.66 & 35.30 & 53.11 & 61.55 &    39.59    &  55.45  & 46.39 & 1.44× &0.58  \\
LazyFormer& 38.90 & 54.41 & 38.71 & 37.18 & 57.61 &    42.23     &      50.66    & 45.67 & 1.38× &0.70  \\
% \midrule 
\name-f & 35.03 & 55.07 & 40.28 & 54.51 & 50.22 &   54.75   &     64.52  & 50.63  & 1.15× &1.43  \\
% PromptCache &  -- & -- & -- & -- & -- & -- & --  & -- & -- &--  \\
% \midrule
\textbf{\name} &  62.66 &  72.23 &  61.85 &  73.32 &  81.59&    74.66   &     65.89   &  70.31    & 1.20× &0.09  \\

\bottomrule

\end{tabular}
}
\label{tab:main_results}
\end{table*}

We evaluate \name on a server equipped with two sockets, each with 24-core Intel(R) Xeon(R) Silver 4410Y processors. The platform provides 512 GB DRAM and a 14 TB hard disk drive (HDD), with the DRAM used to store the attention map database and feature vector database. In addition, the platform includes an NVIDIA A100 GPU with 80 GB of HBM. Since AttnCache performs inference only once and generates sentence embeddings solely during the prefill phase—without creating or storing KV cache—a 80 GB GPU is sufficient for small models, such as Llama-3-3B. If the GPU runs out of memory, online inference can only be performed on the CPU. To build the feature vector database, we use Faiss \cite{johnson2019billion}, a vector database enabling efficient similarity search by the Hierarchical Navigable Small Worlds algorithm \cite{malkov2018efficient}. We use the standard LLM inference as the baseline, named \textit{full model.}

% We use the original transformer model as the standard baseline, named \textit{full model.}

\subsection{Details of Implementation}
% We use test samples in STS datasets to evaluate \name. We apply 8-bit (int8) quantization to LLMs weights in order to save memory space. 
For each task, we collect the input hidden states and their corresponding attention maps at each layer, which are used for training the feature projector and building databases; then we randomly select 1K samples that are not involved in training to measure \name.  
% Across the seven datasets, we build the attention map database using 7K sentences and use 7K test samples for speed measurement. 
The dimensions of the feature vector and batch size are set to 128 and 64, respectively. To maintain high inference accuracy, we set the similarity threshold $\theta$ to 0.99, and set $\alpha$, which is used to train the feature projectors (see Equation~\ref{eq:labels_for_feature_projector}), to 0.2. For efficient similarity search, we construct the feature vector database using Faiss~\cite{johnson2019billion}. Faiss is highly efficient for similarity search. For example, our evaluation shows that with Faiss, searching 100K vectors with a vector-dimension size of 128 takes less than 0.5 ms, which yields 360$\times$ and 10$\times$ speedups over self-attention computation and embedding generation, respectively. 
As a result, the search process does not create a performance bottleneck for \name. In addition, we store each layer's attention maps as a file object in memory. When retrieving attention maps as a batch, the file objects are mapped into a contiguous virtual memory space as a tensor without a memory copy. After self-attention calculation, the file objects
are unmapped. When the combined size of the attention map and feature vector databases exceeds the available DRAM on our platform, we evaluate model performance using a hybrid storage setup that spans DRAM and HDD. While this configuration introduces I/O latency during access, it does not affect the correctness or quality of the model outputs. 
In this case, to evaluate the model inference time on a ``virtual'' big DRAM system with enough capacity to store attention maps, we use our limited DRAM assuming that the needed attention maps are in the DRAM for measuring time.

% , and assume that both the attention map database and feature vector database are memory-resident in DRAM. 

% In such a case, the search overhead (excluding the feature projector overhead) is ignored, because the search time on the DRAM is only a small portion (at most 2\%) of total inference time due to highly optimized Faiss. 

% We implement \name with PyTorch 2.5.1. In the evaluation, all 48 CPU cores are fully utilized for maximum thread-level parallelism to minimize inference time.

\subsection{Baselines}
%We choose three attention reuse approaches as baselines.

\begin{table*}[t]
    \centering
    \renewcommand{\arraystretch}{0.9}
    % \scriptsize
    \tiny 
    \caption{MMLU Math performance on Llama-3-3B under different n-shot settings.}
 \label{tab:mmlu_math_n_shot}
 \resizebox{\textwidth}{!}{ 
    \begin{tabular}{
        ccccccccc
    }
    \toprule
    \multirow{2}{*}{\textbf{Type}} & 
    \multicolumn{2}{c}{Context Length} & 
    \multicolumn{2}{c}{Accuracy} &  
    \multicolumn{2}{c}{Attn Speedup}&
    \multicolumn{2}{c}{ E2E Speedup}\\
    
    \cmidrule(lr){2-3} \cmidrule(lr){4-5} \cmidrule(lr){6-7} \cmidrule(lr){8-9}
    & Max Len.  & Avg. Len.  & Baseline & AttnCache
    & CPU & GPU & CPU & GPU \\
    
    \midrule
    
    0-shot & 373  & 93.85 & 28.70  & 29.57  & 2.09x & 3.01x & 1.22x & 1.66x \\
    1-shot & 483 & 178.86 & 40.87 & 38.26  & 2.04x & 2.99x & 1.23x & 1.62x \\
    2-shot & 613 & 284.79 &  41.74 & 42.61  & 2.11x & 2.97x & 1.19x & 1.64x\\
    3-shot & 784 & 393.12 & 40.00 & 39.13  & 2.13x & 3.01x & 1.27x & 1.62x \\
    4-shot & 954 & 490.28 & 43.48 & 41.74  & 2.07x & 3.02x & 1.25x & 1.63x \\
    5-shot & 1019 & 560.28 & 47.83 & 46.96  & 2.12x & 3.00x & 1.24x & 1.65x \\
    
    \bottomrule
    \end{tabular}
    } % end resizebox
\end{table*}

We use three baselines for evaluation.

\textbf{LazyFormer} \cite{ying2021lazyformer}  divides all layers of the transformer to multiple subblocks. In each subblock, the attention maps are only computed in the first layer and then used by the remaining layers in the same subblock. Like LazyFormer, we set the number of layers in each sub-block to 2. %%%%meaning that for every two layers, the attention maps of the first layer are computed and used by the second layer.

\textbf{SAN} \cite{xiao2019sharing} shares attention maps across multiple adjacent layers. But different from Lazyformer, SAN does not use a uniform subblock size (i.e., the number of transformer layers in a subblock). The subblock size is dynamically determined based on the similarity of layers in terms of the JS divergence \cite{menendez1997jensen}. 

%\paragraph{\name-f} \cite{feng2023\name}: \name-f utilizes the similarity of attentin maps in the same layer between different input sentences to reuse the attentin maps. It needs to map hidden states to embeddings and do a search at each layer to find similar attention maps.

\textbf{\name-f} is a variant of \name. \name-f applies memoization at the transformer layer level instead of the whole model level (as \name does). In particular, at each layer, \name-f searches the attention map database for similar attention maps, hence applying a \textit{f}ine-grained memoization.  Moreover, \name-f does not consider sequence length when training the feature projector, meaning that the \( y \) in Equation \ref{eq:labels_for_feature_projector} does not take into account the computation of \( ||s_{1}- s_{2}||_{1} \).
%\name-f utilizes the similarity of attentin maps in the same layer between different input sentences to reuse the attentin maps. It needs to map hidden states to embeddings and do a search at each layer to find similar attention maps.

%We use test samples in STS datasets to evaluate \name. Since \name is orthogonal to model quantization shown in Section \ref{orthogonal}, we employ 8-bit (int 8) quantization in order to save memory space. We collect the attention maps and input embeddings of 1,000 sentences for each task to train feature projector and build databases, and use 1000 samples to measure the inference time of self attention and remaining samples are used for testing.  In total, we build the attention map database using 7000 sentences and use 7000 test samples for the speed measurements. To maintain a high accuracy, we set the similarity threshold of the feature vectors to 0.99 and set $\alpha$ to 0.2. In addition,  we use the Speedup Degradation Ratio $\gamma$ metric  \cite{he2024matters} to quantify the trade-off between speed and performance degradation.

% We use the Speedup Degradation Ratio  \cite{he2024matters} $\gamma$ to quantify trade-off between speed and performance degradation.

To quantify the trade-off between speed and performance degradation, we adopt the Speedup Degradation Ratio $\gamma$ \cite{he2024matters} as an evaluation metric.

\begin{equation}
\gamma =  
\frac{\text{Avg}_{\text{full}} - \text{Avg}_{\text{method}}}{\text{Speedup}_{\text{method}} - \text{Speedup}_{\text{full}}}
\label{eq:gamma}
\end{equation}

% {
% \setlength{\abovedisplayskip}{1pt}  
% \setlength{\belowdisplayskip}{1pt}  
% \setlength{\abovedisplayshortskip}{0pt}
% \setlength{\belowdisplayshortskip}{6pt}
% \begin{equation}
% \gamma =  
% \frac{\text{Avg}_{\text{full}} - \text{Avg}_{\text{method}}}{\text{Speedup}_{\text{method}} - \text{Speedup}_{\text{full}}}
% \label{eq:gamma}
% \end{equation}
% }

where $\text{Avg}_{full}$ and $\text{Avg}_{method}$ are the average performance of LLMs and each method across the seven  tasks respectively, and  $\text{Speedup}_{full}$ and $\text{Speedup}_{method}$ represent the corresponding speedup respectively. %$\gamma$ measures the degree of performance degradation associated with 1\% enhancement in speedup and 
A smaller $\gamma$ indicates that the method is more efficient.

\subsection{Main Results}

Table \ref{tab:main_results} summarizes the experimental results on the STS datasets.  Across various models (Llama2-7B, Llama3-8B, and Mistral-7B), SAN and LazyFormer both lead to notable performance declines, despite achieving higher speedups. For instance, LazyFormer results in an average 25.39\% performance decline (from 71.88\% to 46.49\%) for Llama-3-8B, with a speedup of 1.42$\times$, corresponding to a $\gamma$ of 0.60. We also notice that the inter-sentence methods (i.e. \name-f and \name) exhibit higher performance but lower speedup compared to intra-sentence methods because they only reuse attention maps with high similarity. For example, for Llama-2-7B, \name-f and \name achieve average performance of 50.39\% and 67.75\% with corresponding speedups of 1.14$\times$ and 1.19$\times$, while SAN and LazyFormer yield 30.34\% and 34.48\% performance with speedups of 1.39$\times$ and 1.45$\times$ separately. 
Moreover, \name maintains near full model performance on various datasets and strikes a better balance between speed and performance, with $\gamma$ values of 0.04, 0.11 and 0.09 for three LLMs, making it a superior method for the acceleration of self attention.

% \name consistently outperforms \name-f across all tasks. %That is because \name-f does not take into account the impact of sequence length, which leads to situations where the replaced attention maps have high similarity but a large difference in token sequence lengths. This situation introduces errors that propagate across layers, degrading performance. 
% The key limitation of \name-f is its failure to account for sequence length, leading to cases where similar attention maps are reused despite significant token length differences. This mismatch introduces errors that accumulate across layers, degrading performance.

 As shown in Table \ref{tab:time_breakdown}, \name-f performs embedding and vector search at each layer, even when no reusable attention maps are found, increasing latency. 
In contrast, \name performs this computation only once at the beginning of inference to determine whether to reuse attention maps, eliminating embedding and vector search overhead during subsequent layers. Consequently, \name achieves a higher $\gamma$ than \name-f. As illustrated in Figure~\ref{fig:mmlu_stem_small}, AttnCache speeds up Llama-3-3B on MMLU STEM, achieving up to 2×/3× attention and 1.2×/1.6× end-to-end speedups on CPU/GPU, with minimal accuracy loss. Similar results are observed under varying context lengths by adjusting the n-shot input settings, as shown in Table~\ref{tab:mmlu_math_n_shot}.

\subsection{Evaluation on Dense and MoE Language Models.}

\begin{table}
\centering
\caption{\textbf{Time (ms) breakdown for \name-f and \name in a layer of Llama-3-8B.}}
\resizebox{\linewidth}{!}{ 
\begin{tabular}{cccc}
\toprule
\textbf{Time (ms)} & \textbf{Full Model} & \textbf{\name-f} & \textbf{\name} \\
\midrule
Embedding          & N/A                      &   32      & N/A                      \\
Vector Searching          & N/A                  & 2      & N/A                         \\
APM Fetching          & N/A                      & 18      & N/A                        \\
Q Computation   & 73                        &  N/A        & N/A                   \\
K Computation   & 41                       &  N/A         & N/A                   \\
Rotary Pos Encoding & 124                        &  N/A    & N/A                      \\
V Computation   & 41                      & 41           & 41                  \\
AM Computation & 88                    & N/A           & N/A                \\
Other (e.g. AM $\bullet$ V)  & 115                     & 115        & 115                     \\
\midrule
\textbf{Attention}               &482                      &  208     & 156                       \\
\midrule
\textbf{FFN}              & 830                     & 830     & 830                                        \\
\midrule
\textbf{Total}              &1312                    &  1038   & 986                           \\

\bottomrule
\label{tab:time_breakdown}
\end{tabular}}
\end{table}

To fit the LLMs into a single GPU for MoE models, we use bitsandbytes \cite{bitsandbytes2025} quantization to reduce the GPU memory footprint of LLMs. Specifically, for Llama-3-8B, Deepseek-MoE-16B, and Qwen1.5-MoE-A2.7B, we apply NF4 (Normal Float 4) quantization \cite{dettmers2023qlora}. The experimental results are presented in Table~\ref{tab:dense_and_moe_perfermance}. AttnCache achieves up to a 2.43$\times$ attention speedup and a 1.48$\times$ end-to-end speedup with only minor accuracy degradation, demonstrating the robustness of the method. These results confirm that AttnCache is both effective and generalizable across Dense and MoE model architectures.

\begin{table}[h]
\centering
\caption{MMLU Math performance on Dense and MoE models.}
\resizebox{\columnwidth}{!}{%
\begin{threeparttable}

\begin{tabular}{lccc}
\toprule
\textbf{Models} & \textbf{Llama-3-8B} & \textbf{Deepseek-MoE-16B} & \textbf{Qwen1.5-MoE-A2.7B} \\
\midrule
\# Parameters & 8.03B & 16.40B & 14.30B \\
Model Type & Dense & MoE & MoE \\
Quant Type & NF4  & NF4 & NF4 \\
Memory Footprint & \textasciitilde
7G & \textasciitilde
10G & \textasciitilde
9G \\
\midrule
Baseline Accuracy & 41.74 & 30.43 & 35.65\\
AttnCache Accuracy & 40.00 & 28.70 & 34.78\\
\midrule

Attn Speedup &1.67x &2.43x  & 2.28x \\
E2E Speedup  &1.48x  &1.12x & 1.15x \\
\bottomrule
\end{tabular}
% \begin{tablenotes}
%         \footnotesize
%         \item[\dag] \textcolor{blue}{Normal Float 4.}
% \end{tablenotes}

\end{threeparttable}

}
\label{tab:dense_and_moe_perfermance}

\end{table}

\subsection{Evaluation on Transformer Encoder Models.}
We evaluate the effectiveness of AttnCache on Transformer Encoder models (BERT base, RoBERTa, and DeBERTa) on the SST-2 dataset. We collect the hidden states and attention maps from the SST-2 training set to train the feature projector, store the attention maps in the database, and test on the validation set. We set the thresholds for Conservative, Moderate, and Aggressive to 0.995, 0.99, and 0.95, respectively. As shown in Table~\ref{tab:bert_perf}, AttnCache yields an average 1.21$\times$ inference speedup with a modest performance degradation of 1.3\% to 2.7\%. Notably, while some accuracy drop is observed (e.g., in BERT base and RoBERTa), DeBERTa unexpectedly shows slight improvements under conservative reuse, suggesting that cached attention maps can, in some cases, enhance representation quality. These findings confirm that with carefully chosen similarity thresholds, AttnCache can balance efficiency and accuracy even in full-attention encoder settings.

\begin{table}[!t]
\centering
\caption{Model size and architecture type.}
\resizebox{0.85\columnwidth}{!}{%
\begin{tabular}{lccc}
\toprule
\textbf{Models} & \textbf{BERT base} & \textbf{RoBERTa} & \textbf{DeBERTa} \\
\midrule
\# Parameters & 110M & 125M & 139M \\
Model Type & Encoder & Encoder & Encoder \\
\bottomrule
\end{tabular}
}
\label{tab:model-info}
\end{table}

\begin{table}[!t]
\centering
\caption{Accuracy and speedup under different thresholds on SST-2.}
\resizebox{\columnwidth}{!}{%
\begin{tabular}{lccccc}
\toprule
\textbf{\%} & \textbf{Baseline} & \textbf{Conservative} & \textbf{Moderate} & \textbf{Aggressive} & \textbf{Avg. Diff.} \\
\midrule
\textbf{BERT base} & 91.3 & 91.1 & 90.2 & 85.7 & -2.3 \\
\textbf{RoBERTa} & 94.8 & 93.2 & 92.6 & 90.4 & -2.7 \\
\textbf{DeBERTa} & 95.0 & 95.5 & 95.2 & 90.5 & -1.3 \\
\textbf{E2E Speedup} & N/A & 1.10x & 1.18x & 1.34x & 1.21x \\
\bottomrule
\end{tabular}
}
\label{tab:bert_perf}
\end{table}

\section{Analysis}

% \begin{table*}[t]
%     \centering
%     % \renewcommand{\arraystretch}{1}
%     \scriptsize
%     \caption{MMLU Math performance on Llama-3-3B under different n-shot settings.}
%  \label{tab:mmlu_math_n_shot}
%  \resizebox{\textwidth}{!}{ 
%     \begin{tabular}{
%         ccccccccc
%     }
%     \toprule
%     \multirow{2}{*}{\textbf{Type}} & 
%     \multicolumn{2}{c}{\textbf{Context Length}} & 
%     \multicolumn{2}{c}{\textbf{Accuracy}} &  
%     \multicolumn{2}{c}{\textbf{Attn Speedup}}&
%     \multicolumn{2}{c}{\textbf{ E2E Speedup}}\\
    
%     \cmidrule(lr){2-3} \cmidrule(lr){4-5} \cmidrule(lr){6-7} \cmidrule(lr){8-9}
%     & \textbf{Max Len. } & \textbf{Avg. Len.}  & \textbf{Baseline} & \textbf{AttnCache}
%     & \textbf{CPU} & \textbf{GPU} & \textbf{CPU} & \textbf{GPU} \\
    
%     \midrule
    
%     0-shot & 373  & 93.85 & 28.70  & 29.57  & 2.09x & 3.01x & 1.22x & 1.66x \\
%     1-shot & 483 & 178.86 & 40.87 & 38.26  & 2.04x & 2.99x & 1.23x & 1.62x \\
%     2-shot & 613 & 284.79 &  41.74 & 42.61  & 2.11x & 2.97x & 1.19x & 1.64x\\
%     3-shot & 784 & 393.12 & 40.00 & 39.13  & 2.13x & 3.01x & 1.27x & 1.62x \\
%     4-shot & 954 & 490.28 & 43.48 & 41.74  & 2.07x & 3.02x & 1.25x & 1.63x \\
%     5-shot & 1019 & 560.28 & 47.83 & 46.96  & 2.12x & 3.00x & 1.24x & 1.65x \\
    
%     \bottomrule
%     \end{tabular}
%     } % end resizebox
% \end{table*}

\subsection{Impacts of Model Quantinization and Pruning}\label{orthogonal}

\begin{table}[!t]
\centering
\caption
{
\textbf{Integration with model Quantization and Pruning}. “w/Quant”, "w/AttnDrop" and "w/BlockDrop" denotes integration with the quantized model, attention pruning and layer pruning repectively. }
\resizebox{\linewidth}{!}{ 
\begin{tabular}{l|ccccccc|c}
\toprule 
\multicolumn{6}{c}{
Llama-3.2-3B} \\
\midrule
 \textbf{Method}  & STS13 & STS14 & STS15 & STS16 & Avg. \\
\midrule
  Full Model  & 76.56 & 60.05 & 74.76 & 79.30 & 72.67\\
\name &  74.74 & 59.95 & 74.19 & 77.38 & 71.57\\
 
\midrule 
Quanto & 75.27 & 57.55 & 74.41 & 76.96 & 71.05 \\
w/Quanto & 74.25 & 54.75 &  74.49 &  76.92 & 70.10 \\
AttnDrop & 75.33 & 59.04 & 69.92 & 78.37 & 70.67\\
w/AttnDrop  & 73.21 & 56.01 & 69.48 & 75.49 & 68.55\\
BlockDrop & 67.98 & 50.44 & 72.42 & 75.52 & 66.59 \\
w/BlockDrop & 67.18 & 50.49 & 70.44 & 73.67 & 65.45 \\

\bottomrule

\end{tabular}
}
\label{tab:integration} 
\end{table}

Model quantization represents weights and activations with lower-precision data type, and can improve efficiency in memory usage and inference speed. We integrate \name with quantization and apply Quanto \cite{Optimum-quanto} to all weights, and use 4-bit quantization. We also combine \name with recent LLM pruning methods, AttnDrop and BlockDrop \cite{he2024matters}, which remove redundant attentions and layers by measuring the similarity between input and output of each layer. Table \ref{tab:integration} shows the results. The integration of model quantization and pruning with \name maintains performance: the difference between \name and Quanto/BlockDrop is only 1\%, and the difference between \name and AttnDrop is only 2\%, on average.

% \subsection{Impacts of Database Size}

\subsection{Impact of Similarity Thresholds}

\begin{figure}[!t]
	\centering
	\includegraphics[width=0.9\linewidth]{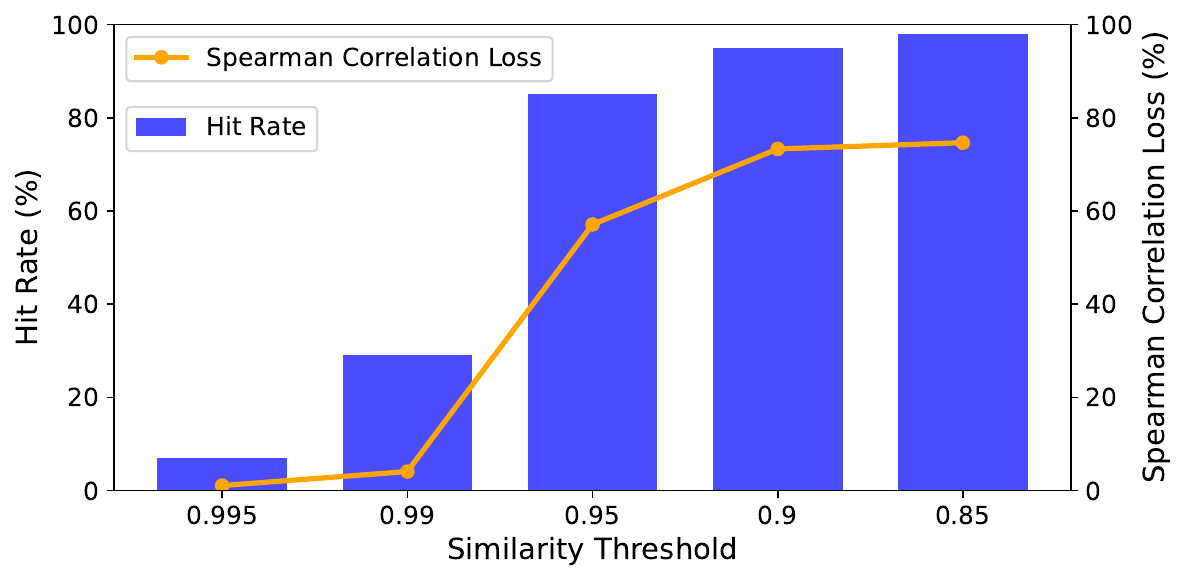}
	\caption{Impact of Threshold on Spearman Correlation.}
\label{fig:threshold_impact}
\end{figure}

Assume that there are $N$ input sentences for an LLM to generate sentence embeddings,  we count how many times \name is successfully applied (indicating similar attention maps are found), denoted as $M$. We use the ratio $ M/N$ as the hit rate. We randomly select 100 sentences from STS15, and change the similarity threshold $\theta$ from 0.995 to 0.85. We measure the hit rate and loss in the Spearman correlation score. As shown in Figure \ref{fig:threshold_impact}. When we reduce $\theta$, the hit rate increases, which means that more attention maps are found and \name leads to  higher acceleration. However, this might lead to replacement with less similarity, decreasing the performance. By setting $\theta$ to 0.99, our results show that \name provides 30\% hit rate with only 2\% reduction in the Spearman correlation score.

%As a preliminary study, we use a similarity threshold to control whether \name should be applied. We use the search results from  the feature vector database to prefetch attention maps only when the similarity is larger than or equal to the threshold. The feature vector database uses Euclidean distance for searching, and the feature vectors are normalized before being stored. Since sentences from the same dataset have very similar lengths, it is always possible to find a feature vector with a threshold greater than 0.9 (Euclidean distance less than 0.1) for any query request. We randomly selected 100 sentences from STS15 and change the similarity threshold from 1 to 0.85 and measure the hit rate and spearman correlation score loss. As shown in Figure \ref{fig:threshold_impact},  when we reduce the threshold, the hit rate increases, which means more reusable attention maps are found and achieves higher acceleration. However, this might lead to some attention maps being substituted with ones that are less similar, decreasing the spearman correlation scroes. Thus thresholds are very important in balancing the tradeoff between inference time and inference accuracy. For example, by setting the thresholds as 0.99, our results show that \name provides 30\% hit rate with only 2\% reduction of spearman correlation.

\section{Conclusions}
% The emerging big memory system brings new optimization opportunities for accelerating LLMs. 
In this paper, we propose \name to accelerate self attention inference during the prefill stage of LLM inference. Our work is based on the observation that semantically different input sentences can exhibit highly similar attention maps across layers or heads during inference computation. 
By pre-storing similar attention maps in a database, when generating a new sentence embedding, the most similar attention map can be retrieved from the attention map database and reused to reduce self-attention computation.
\name provides an average 1.2$\times$ end-to-end and 2$\times$ attention speedup on CPU, and 1.6$\times$ end-to-end and 3$\times$ attention speedup on GPU, with negligible accuracy loss. 

%Our key insight is that semantically different input sentences can exhibit highly similar attention maps across layers or heads. By storing these maps in a database, \name retrieves and reuses them via a memory cache, reducing redundant self-attention computations. This approach achieves an average $1.2\times$ speedup with minimal performance loss."

% \bibliography{mlsys2025style/example_paper, mlsys2025style/li}

% \bibliography{example_paper}
\bibliography{mlsys2025style/main}
\bibliographystyle{mlsys2025style/mlsys2025}

%%%%%%%%%%%%%%%%%%%%%%%%%%%%%%%%%%%%%%%%%%%%%%%%%%%%%%%%%%%%%%%%%%%%%%%%%%%%%%%
%%%%%%%%%%%%%%%%%%%%%%%%%%%%%%%%%%%%%%%%%%%%%%%%%%%%%%%%%%%%%%%%%%%%%%%%%%%%%%%
% SUPPLEMENTAL CONTENT AS APPENDIX AFTER REFERENCES
%%%%%%%%%%%%%%%%%%%%%%%%%%%%%%%%%%%%%%%%%%%%%%%%%%%%%%%%%%%%%%%%%%%%%%%%%%%%%%%
%%%%%%%%%%%%%%%%%%%%%%%%%%%%%%%%%%%%%%%%%%%%%%%%%%%%%%%%%%%%%%%%%%%%%%%%%%%%%%%
\appendix

% \section{Please add supplemental material as appendix here}
% %
% Put anything that you might normally include after the references as an appendix here, {\it not in a separate supplementary file}. Upload your final camera-ready as a single pdf, including all appendices.

% %%%%%%%%%%%%%%%%%%%%%%%%%%%%%%%%%%%%%%%%%%%%%%%%%%%%%%%%%%%%%%%%%%%%%%%%%%%%%%%
% %%%%%%%%%%%%%%%%%%%%%%%%%%%%%%%%%%%%%%%%%%%%%%%%%%%%%%%%%%%%%%%%%%%%%%%%%%%%%%%

\end{document}